\documentclass[letterpaper, 10 pt, conference]{ieeeconf}  % Comment this line out if you need a4paper

\IEEEoverridecommandlockouts                              % This command is only needed if 
\usepackage{graphicx} % for pdf, bitmapped graphics files
\usepackage{amsmath} % assumes amsmath package installed
\usepackage{multirow}
\usepackage[ruled,vlined]{algorithm2e}
\usepackage{todonotes}
\usepackage{hyperref}
\usepackage{xspace}
\usepackage{footnote}
\makesavenoteenv{tabular}
\makesavenoteenv{table}

\makeatletter
\DeclareRobustCommand\onedot{\futurelet\@let@token\@onedot}
\def\@onedot{\ifx\@let@token.\else.\null\fi\xspace}
\def\eg{\emph{e.g}\onedot} 
\def\ie{\emph{i.e}\onedot} 
 
\def\etc{\emph{etc}\onedot} \def\vs{\emph{vs}\onedot}
 
\def\etal{\emph{et al}\onedot}
\makeatother

\title{\LARGE \bf
Semantic Foreground Inpainting from Weak Supervision
}

\author{Chenyang Lu and Gijs Dubbelman
	\thanks{
		This work was supported by the Netherlands Organization for Scientific Research (NWO) in the context of the i-CAVE project.} %Use only for final RAL version
	\thanks{Authors are with the Mobile Perception Systems research lab of the SPS/VCA group, Department of Electrical Engineering,  Eindhoven University of Technology, The Netherlands.
		{\tt\footnotesize \{c.lu.2, g.dubbelman\}@tue.nl}}%
	}

\begin{document}

\maketitle
\thispagestyle{empty}
\pagestyle{empty}

%%%%%%%%%%%%%%%%%%%%%%%%%%%%%%%%%%%%%%%%%%%%%%%%%%%%%%%%%%%%%%%%%%%%%%%%%%%%%%%%
\begin{abstract} 
	
	Semantic scene understanding is an essential task for self-driving vehicles and mobile robots. In our work, we aim to estimate a semantic segmentation map, in which the foreground objects are removed and semantically inpainted with background classes, from a single RGB image. This \textit{semantic foreground inpainting} task is performed by a single-stage convolutional neural network (CNN) that contains our novel \textit{max-pooling as inpainting} (MPI) module, which is trained with weak supervision, \ie, it does not require manual background annotations for the foreground regions to be inpainted. Our approach is inherently more efficient than the previous two-stage state-of-the-art method, and outperforms it by a margin of 3\% IoU for the inpainted foreground regions on Cityscapes. The performance margin increases to 6\% IoU, when tested on the unseen KITTI dataset. The code and the manually annotated datasets for testing are shared with the research community at \url{https://github.com/Chenyang-Lu/semantic-foreground-inpainting}.
	
\end{abstract}

\section{Introduction}
\label{sect_introduction}

Semantic segmentation \cite{badrinarayanan_segnet:_2017, chen_deeplab:_2018, meletis_training_2018, shelhamer_fully_2017, zhao_pyramid_2017} has been well investigated in recent years with the advances of deep convolutional neural networks (CNNs). The traditional semantic segmentation task is formulated as assigning a specific pre-defined semantic label to each pixel in an image, which is performed solely on the 2-D image domain. Arguably, vanilla semantic segmentation on the 2-D image is insufficient for the understanding of 3-D scenes. For example, for autonomous driving, foreground objects such as vehicles and pedestrians can occlude the road, thus inhibit the reasoning of the complete road layout and further navigation and planning tasks. 

Intuitively, humans are able to reason and obtain more information than CNNs from a single image. For example, if a vehicle is observed, one can still reason that the road exists under and behind that vehicle, even though these regions are occluded. The ability of occlusion reasoning can evidently improve the intelligence of autonomous agents, while it cannot be easily realized by a canonical semantic segmentation CNN that is trained with strong pixel-wise supervision. This is mainly because: 1) a large number of foreground removed ground truth samples would be required for supervised training, and 2) the manual annotation of the occluded regions is a challenging task especially when the scene is complicated and the occluded regions are relatively large.

\begin{figure}[!tbp]
	\centering
	\includegraphics[width=0.9\linewidth]{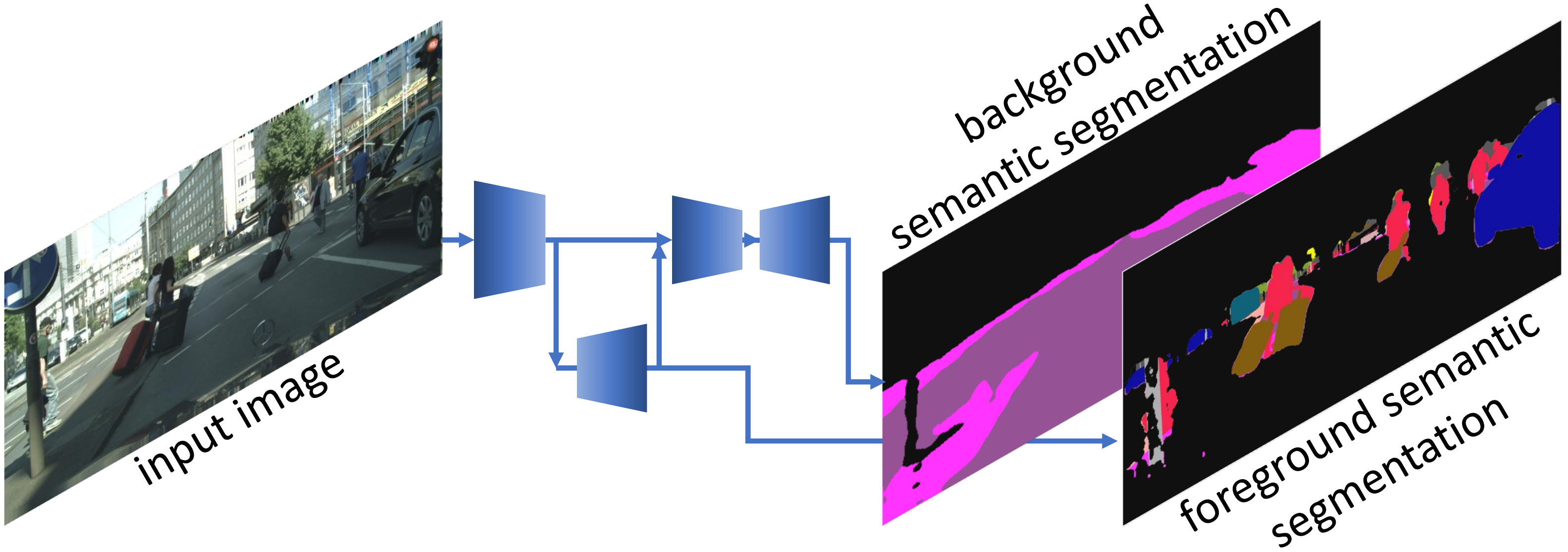}
	\vspace{-10pt}
	\caption{System overview. Our network is able to segment out the foreground objects (binary or optionally multi-class), and simultaneously inpaint the semantic scene behind these foreground objects  with background classes (\textit{road}, \textit{sidewalk}, and \textit{other rigid world}).}
	\label{fig_overview}
	\vspace{-10pt}
\end{figure}

The most recent and relevant work focusing on \textit{semantic foreground inpainting}, \ie, separating and removing foreground objects from the background and semantically inpainting with background classes, is \cite{schulter_learning_2018}. It performs foreground removal in the RGB image domain as a coarse pre-processing step, using automatically generated random rectangular masks to construct weak supervision at occluded regions. This method is detailed in Section~\ref{sect_related_work} as one of the baseline methods. In contrast, our approach, introduced in Section~\ref{sect_methodology}, utilizes the \textit{novel max-pooling as inpainting} (MPI) module, which enables explicit nearby background feature inpainting and facilitates complementary weak supervision without the need for human annotations. In Section~\ref{sect_experiments}, we perform extensive experiments on Cityscapes \cite{cordts_cityscapes_2016} and KITTI \cite{geiger_vision_2013} datasets using our manually annotated test samples, which are shared with the research community.

To summarize, we make the following contributions:

\begin{itemize}
	\item A novel and efficient architecture, for the \textit{semantic foreground inpainting} task, whose performance surpasses state-of-the-art.
	\item Our novel \textit{max-pooling as inpainting} (MPI) module for improving the foreground semantic inpainting task, which can be inserted into any CNN without blocking the gradient flow during training.
	\item We publicly release our manually annotated test samples for the Cityscapes \cite{cordts_cityscapes_2016} and KITTI \cite{geiger_vision_2013} datasets, to facilitate further research.
\end{itemize}

\begin{figure*}[!tbp]
	\centering
	\includegraphics[width=0.9\linewidth]{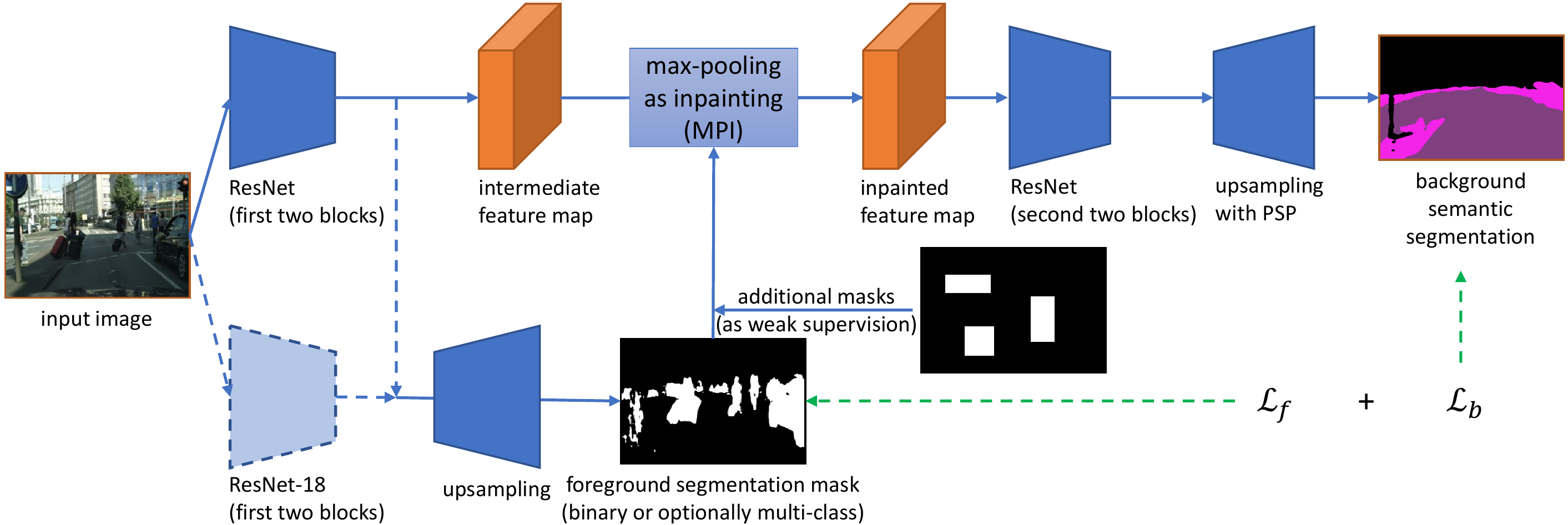}
	%\vspace{-5pt}
	\caption{Network overview. The upper part of the network, which performs the semantic foreground inpainting task, is derived from a state-of-the-art semantic segmentation network, \ie, PSPNet \cite{zhao_pyramid_2017} with the ResNet (-18, -50, \etc) \cite{he_deep_2016} backbone. Note that the backbone is divided into two parts, with a max-pooling as inpainting (MPI) module inserted in the middle. The lower part is a lightweight network (the first two blocks of ResNet-18 with an upsampling module) that segments out the binary foreground objects, which are required by the MPI module in parallel. Please note that this lightweight network can optionally use the features from the upper main branch of the network (indicated by blue dash lines), which results in a more efficient model, and is referred to as \textit{ours-faster}.  Two losses are applied during training. Please see Section~\ref{sect_methodology} for more details.}
	\label{fig_network}
	\vspace{-5pt}
\end{figure*}

\section{Related work}
\label{sect_related_work}

\textbf{Semantic segmentation:} CNNs are widely used in image semantic segmentation \cite{badrinarayanan_segnet:_2017, chen_deeplab:_2018, meletis_training_2018, shelhamer_fully_2017, zhao_pyramid_2017}, and most frameworks are adapted from the fully convolutional network (FCN) \cite{shelhamer_fully_2017}. Segmentation networks often utilize the convolutional feature encoder from backbone networks such as VGG \cite{simonyan_very_2014} and ResNet \cite{he_deep_2016}. Another network architecture named SegNet \cite{badrinarayanan_segnet:_2017} adapts the insight of auto-encoders, with the convolutional feature encoder and decoder being symmetric. Current research also indicates that segmentation can be enhanced by applying conditional random fields (CRFs) as post-processing \cite{chen_deeplab:_2018, zheng_conditional_2015} or using additional adversarial training strategies \cite{luc_semantic_2016, lucic_are_2018}. The difference between the canonical semantic segmentation task and our task is that, besides performing segmentation at background regions, the system needs to hallucinate the semantic background scene behind the foreground objects, which is non-trivial for canonical pixel-wise fully-supervised segmentation approaches.

\textbf{Image inpainting:} Image inpainting \cite{bertalmio_image_2000} aims to recover the missing regions of an image given the surrounding context information. CNNs enable the possibility of image inpainting with large missing areas, as they can extract abstract semantic information from the observable context. The Context Encoder (CE) \cite{pathak_context_2016} network is proposed to inpaint the image with large rectangular areas missing at the image center by applying reconstruction and adversarial loss \cite{goodfellow_generative_2014} in training. CE-like networks \cite{demir_patch-based_2018, iizuka_globally_2017, li_generative_2017, yu_generative_2018, yu_free-form_2018} are proposed with additional discriminative networks applied on locally missing regions or the entire image in a patch-wised manner, which are able to perform inpainting with regions missing at arbitrary positions. The aforementioned approaches are all performed on regular RGB images, and they are trained and tested under the condition that the complete ground truth for the regions to be inpainted is available. However, in our task, the ground truth is not available and the inpainting task should be performed with a domain shift, \ie, from RGB images to semantic maps.

\textbf{Beyond pixel-wise observation:} Less research has been carried out for scene reasoning beyond the observed pixels and behind the foreground objects. In \cite{lu_monocular_2019, lu_hallucinating_2019}, the authors propose a system that transforms front-view images into top-view grids, with the observable grid cells completed using hallucination as the second step. Uittenbogaard \etal \ \cite{uittenbogaard_privacy_2019} propose to remove and inpaint the moving objects in street-view imagery, using the background from temporal context information in the video. It relies on extra video data and cannot remove movable objects that are static. Towards semantic scene understanding that requires fewer dependencies, Liu \etal\ \cite{liu_building_2016} propose a system to hallucinate a depth map and a semantic map, given an RGB image and a noisy, incomplete depth map, which is able to remove the foreground objects. However, the removal is performed based on the output of a traditional semantic segmentation map, and requires additional depth information and a planar world assumption.

\begin{figure}[!tbp]
	\centering
	\includegraphics[width=0.9\linewidth]{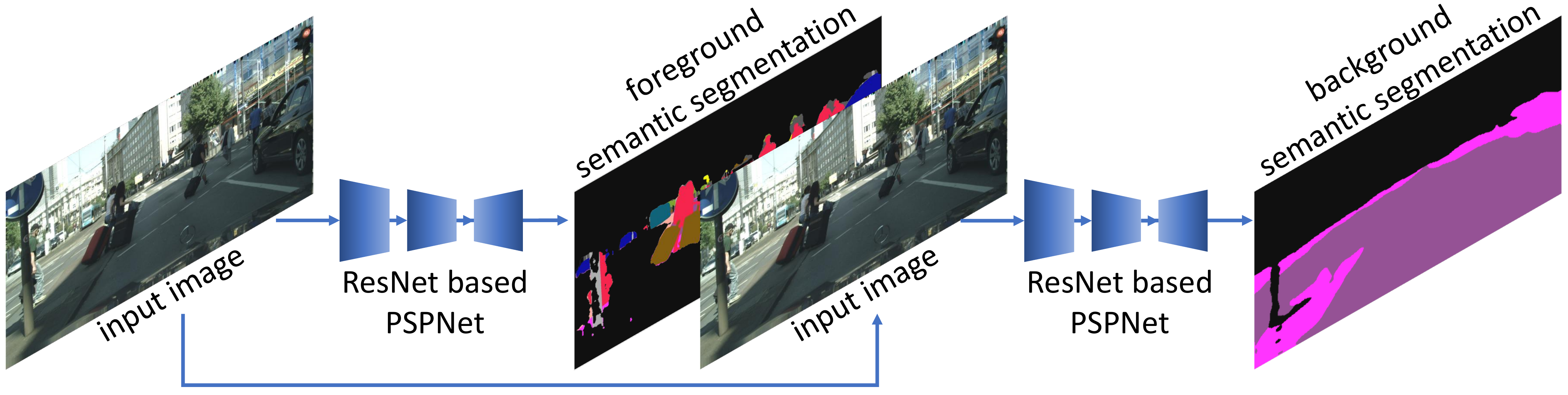}
	\vspace{-10pt}
	\caption{Overview of the baseline \cite{schulter_learning_2018}, which is composed of two stages.}
	\label{fig_baseline}
	\vspace{-10pt}
\end{figure}

Closer to our work, Schulter \etal \ \cite{schulter_learning_2018} propose a CNN to conduct the semantic foreground inpainting task. This method, which is the current state-of-the-art, uses a modified semantic segmentation network, \ie, PSPNet \cite{zhao_pyramid_2017}. It assumes that the foreground objects are always available and are masked out in the corresponding input RGB images. Furthermore, with the ground truth for the masked regions unavailable, a direct supervision on the occluded regions is not possible. Thus, extra \textit{fake} random rectangular foreground masks are generated and applied on the input image along with the \textit{real} foreground masks during training, to provide extra supervision at foreground regions. By creating \textit{fake} foreground masks, the corresponding background ground truth becomes available for supervised training. This approach, which is treated as the main baseline in our work, still requires the foreground masks at the input, resulting in a two-stage pipeline with the first stage computing the foreground masks, see Figure~\ref{fig_baseline}. It can be claimed that a two-stage segmentation and inpainting framework is sub-optimal in terms of efficiency, compared to a single-stage approach like ours.

\section{Methodology}
\label{sect_methodology}

We introduce the proposed method for the \textit{semantic foreground inpainting} task, which is able to infer the semantic scene without foreground objects in a single-stage network, using the novel \textit{max-pooling as inpainting} module.

\subsection{Semantic foreground inpainting}
\label{subsect_semantic_foreground_inpainting}

Given a regular input image, the task is composed of three parts: 1) semantic segmentation of the foreground objects, 2) scene hallucination at the regions which are occluded by the foreground objects, and 3) semantic segmentation of the entire image including the occluded regions. Although semantic segmentation \cite{badrinarayanan_segnet:_2017, chen_deeplab:_2018, meletis_training_2018, shelhamer_fully_2017, zhao_pyramid_2017} and image inpainting \cite{demir_patch-based_2018, iizuka_globally_2017, li_generative_2017, yu_generative_2018, yu_free-form_2018} are well investigated recently, this task is non-trivial and contains several challenges. First, the canonical image inpainting task is solved with the region to be inpainted and its corresponding ground truth available. However in our case, it is not possible to obtain massive ground truth which is occluded by foreground objects. Second, the inpainting task is often solved within the same domain, \eg, the input and output are both RGB images, while our task performs domain transition from RGB images to semantic maps. Third, the inpainting approach usually requires the indication of the regions to be inpainted as input, which results in a complicated two-stage processing pipeline.

\subsection{Single-stage inpainting}

We propose to segment the foreground objects and semantic foreground inpainting simultaneously in parallel, as illustrated in Figure~\ref{fig_network}. The upper part of Figure~\ref{fig_network}, \ie, the main branch of the network, is derived from a canonical semantic segmentation CNN with multiple backbones, \eg, a ResNet-50 \cite{he_deep_2016} based PSPNet \cite{zhao_pyramid_2017} as in \cite{schulter_learning_2018}. Compared to the baseline \cite{schulter_learning_2018}, the differences include: 1) the foreground objects mask is provided at the middle of the network, instead of at the input, and 2) an additional intermediate feature process is performed by the proposed MPI module. The lower part of Figure~\ref{fig_network}, \ie, the auxiliary branch, represents a lightweight CNN which outputs the segmentation of the foreground objects, with the predictions processed in the middle of the main branch. Thus, the foreground segmentation and the early stage of the main network can perform their computations in parallel. Please note the features used for foreground segmentation can also be shared from the main branch, as illustrated with dash lines in Figure~\ref{fig_network}, which could further improve the efficiency with some performance degradation. This variant of our approach, named \textit{ours-faster}, is also evaluated in the experiments.

\begin{algorithm}[!tbp]
	\KwIn{$F_{raw}$: intermediate feature map, $M$: binary foreground mask with 1 being background.} 
	\KwOut{$F_{inpainted}$: inpainted feature map. }
	
	$M$ = max-pooling($M$)\;
	
	$F_{background}$ = $F_{raw} * M$\;
	
	({\footnotesize $*$ denotes element-wise multiplication})
	
	$F_{patch}$ = zero tensor with size same as $F_{background}$\;
	
	$M_{old}$ = $M$\;
	
	\While{$0$ exists in $M_{old}$}{
		$F_{background}$ = max-pooling($F_{background}$)\;
		$M_{new}$ = max-pooling($M_{old}$)\;
		$F_{patch}$ += ($M_{new}$ - $M_{old}$) * $F_{background}$\;
		$M_{old}$ = $M_{new}$\;
	}
	
	$F_{inpainted}$ = $F_{raw}$ * $M$ + $F_{patch}$\;
	\caption{{\bf max-pooling as inpainting (MPI)} \label{algo_MPI}}
\end{algorithm}

\begin{figure}[!tbp]
	\centering
	\vspace{-10pt}
	\includegraphics[width=0.95\linewidth]{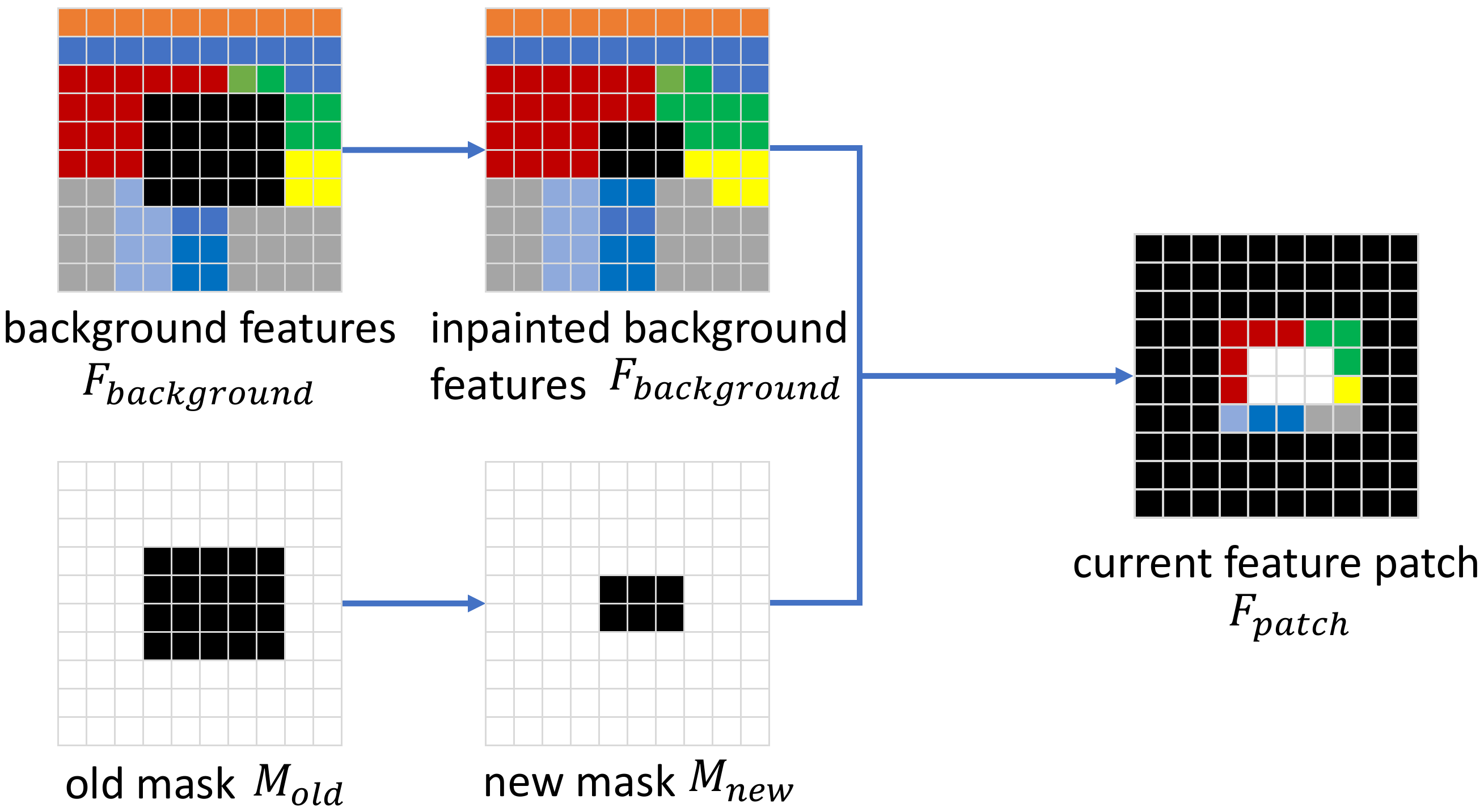}
	\vspace{-10pt}
	\caption{A visualized example of a single iteration in the loop of Algorithm~\ref{algo_MPI}. The binary foreground mask and the background features with foreground region blacked out are simultaneously processed by a canonical max-pooling operation, with kernel size 3, stride 1, and padding 1. The updated features can then be tracked by comparing the difference between the old and new foreground masks, which is referred to as a feature patch. After the final iteration, the patches for each iteration are accumulated and used for the final inpainted feature map.}
	\label{fig_MPI_illustration}
	\vspace{-10pt}
\end{figure}

With this, the semantic foreground inpainting task can be achieved with a single-stage computational efficient network, instead of a pipeline with two separate networks in sequence. Also, the predicted foreground segmentation can be extended with multiple classes, which is not considered in our experiments, as the MPI module requires binary foreground maps and this work is focusing on the quality of the inpainted background.

\begin{figure*}[!tbp]
	\centering
	\includegraphics[width=0.9\linewidth]{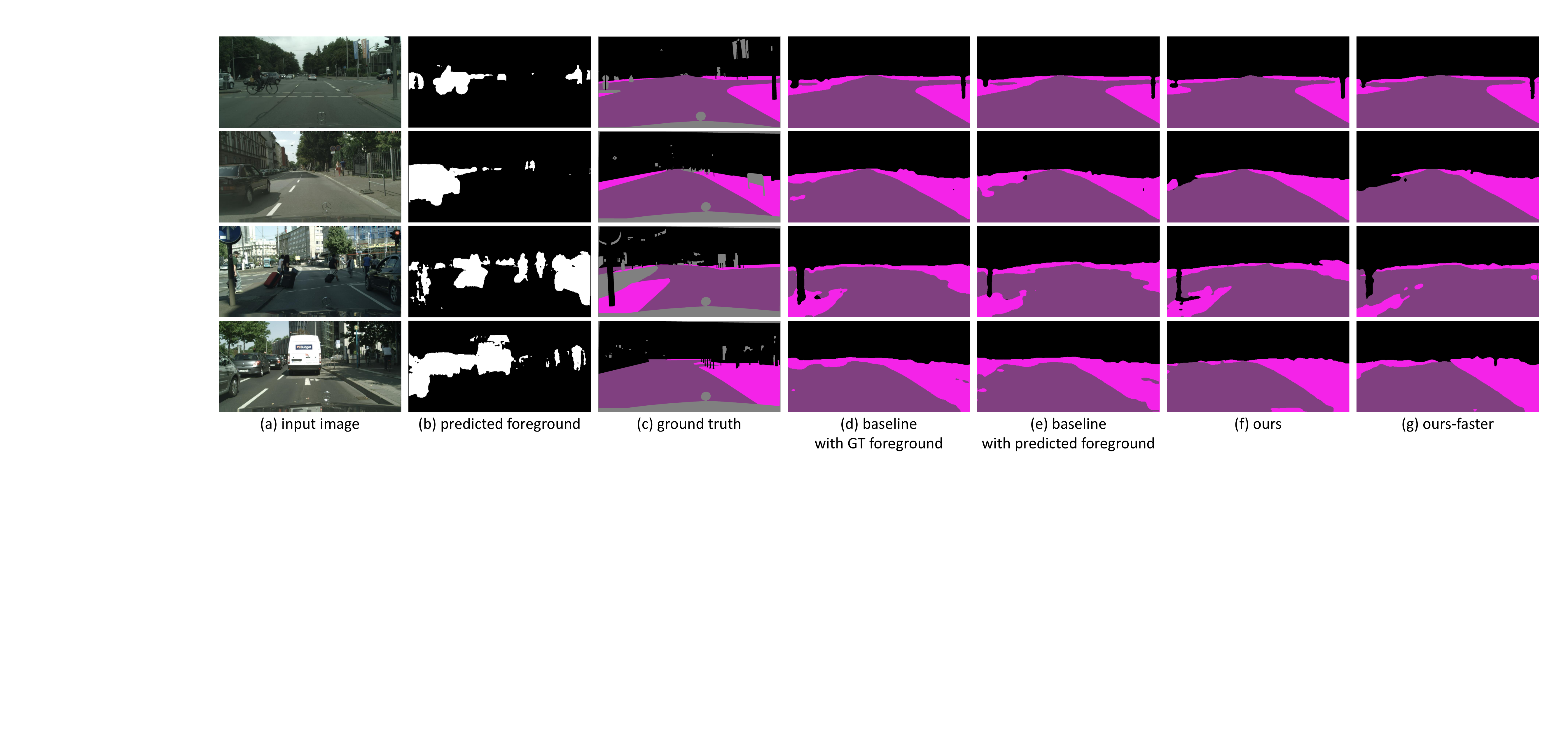}
	\vspace{-10pt}
	\caption{Qualitative results of different methods tested on the Cityscapes dataset. (a) is the input RGB image, (b) is the predicted foreground mask by our network, (c) is the manually annotated segmentation of the background, (d)-(g) are the results of the different methods. }
	\label{fig_main_results}
	\vspace{-5pt}
\end{figure*}

\begin{table*}
	\renewcommand{\arraystretch}{1}
	\caption{Performance of the proposed approach and the baseline.}
	\vspace{-10pt}
	\begin{center}
		\begin{tabular}{c||c|c|c|c|c|c|c}
			\hline
			method & backbone & \# of parameters & inference time (ms) & \# of stages & mask used  &  IoU (all regions)  & IoU (foreground region) \\
			\hline
			\hline
			baseline \cite{schulter_learning_2018}& \multirow{4}{*}{ResNet-18} & 34.8M & 33 &  two  & \textit{ground truth}  & \textit{82.0} & \textit{58.8} \\
			baseline \cite{schulter_learning_2018} &  & 34.8M & 33 & two & predicted  & 81.5 & 57.9 \\
			ours & & 22.5M & 32 & one & predicted  & \textbf{82.3} & \textbf{60.0} \\
			ours-faster & & \textbf{21.8M} & \textbf{22} & one & predicted  & 82.2 & 59.5 \\
			\hline
			baseline \cite{schulter_learning_2018} & \multirow{4}{*}{ResNet-50} & 106.6M & 63 & two & \textit{ground truth} & \textit{82.4} & \textit{58.1} \\
			baseline \cite{schulter_learning_2018} &  & 106.6M & 63 & two & predicted  & 82.2 & 57.8 \\
			ours & & \textbf{58.4M} & 44 & one & predicted & \textbf{83.2} & \textbf{60.8} \\
			ours-faster & & 59.5M\footnotemark  & \textbf{34} & one & predicted & \textbf{83.2} & 60.7 \\
			\hline
		\end{tabular}
	\end{center}
	\label{tab_main_results}
	\vspace{-15pt}
\end{table*}

\subsection{Max-pooling as inpainting}

The key contribution of our work is the \textit{max-pooling as inpainting} module. This module takes an \textit{intermediate feature map} and a corresponding predicted \textit{foreground segmentation mask} as inputs, and outputs a new \textit{inpainted feature map} with the foreground region inpainted with nearby features, as shown in Figure~\ref{fig_network}. The formal definition can be found in Algorithm~\ref{algo_MPI} and a visualization of a single iteration is provided in Figure~\ref{fig_MPI_illustration}.

The overall motivation is to improve and enrich the intermediate features for the task of semantic foreground inpainting, as the intermediate feature map of the raw input image contains the features of the unwanted foreground objects. Inspired by the traditional morphological operations and nearest neighbor inpainting techniques in computer vision, the MPI module aims to explicitly provide extra features at the foreground regions, instead of implicitly learning it as in \cite{schulter_learning_2018}. This module is easy to implement and flexible to use, since it can be inserted into any CNN at any position without blocking the gradient flow, and is constructed using canonical max-pooling operation in an iterative manner, which is available in all neural network frameworks. 

\footnotetext{Foreground upsampling has increased input channels from 128 (ResNet-18)  to 512 (ResNet-50), which results in higher \# of parameters when sharing the feature extractor. Note that the inference time is still reduced as the computational time for foreground feature extractor is no longer needed.}

Specifically, as the preparation of the main iteration loop, as visualized in Figure~\ref{fig_MPI_illustration}, we first black out the features at the foreground regions in the raw intermediate feature maps $F_{raw}$, and obtain a new feature map $F_{background}$ that only contains the background features. Then for each iteration, a conventional max-pooling layer is applied iteratively on the background feature map $F_{background}$, which shifts the nearby background features into the foreground regions. In the meantime, $M$ is pooled simultaneously with the $F_{background}$ and compared with the previous mask $M_{old}$ to index the pixels that are updated in the current iteration. Thus, a new feature map $F_{patch}$ can be maintained over all the iterations. Finally, $F_{patch}$ is merged with the original $F_{raw}$ for an inpainted feature map $F_{inpainted}$, using the index of the original foreground mask $M$. The output of the MPI module, \ie inpainted feature map $F_{inpainted}$, can then be processed further by canonical convolutional operations without blocking the gradient flow at the background regions. Note that, we also apply an extra max-pooling operation on the foreground masks before the main process, to eliminate the effect of features at boundary regions. From the experiments, presented in Section~\ref{sect_experiments}, we empirically found that the MPI module is best inserted after the second block of the used ResNet \cite{he_deep_2016} feature extractor.

\subsection{Weak supervision}
\label{subsect_weak_supervision}

As mentioned in Section~\ref{subsect_semantic_foreground_inpainting}, an important challenge of the semantic foreground inpainting task is that the ground truth of the occluded background is usually not available for training, which means a direct supervision on the inpainted regions cannot be applied. In our approach, the supervision for the inpainted foreground regions is realized in two different manners. The first one is similar to the aforementioned \textit{fake} foreground mask generation of the baseline \cite{schulter_learning_2018}. The only difference is that, in our case, the foreground masks are applied on the intermediate feature maps right before the MPI module. 

\begin{figure*}[!tbp]
	\centering
	\includegraphics[width=0.95\linewidth]{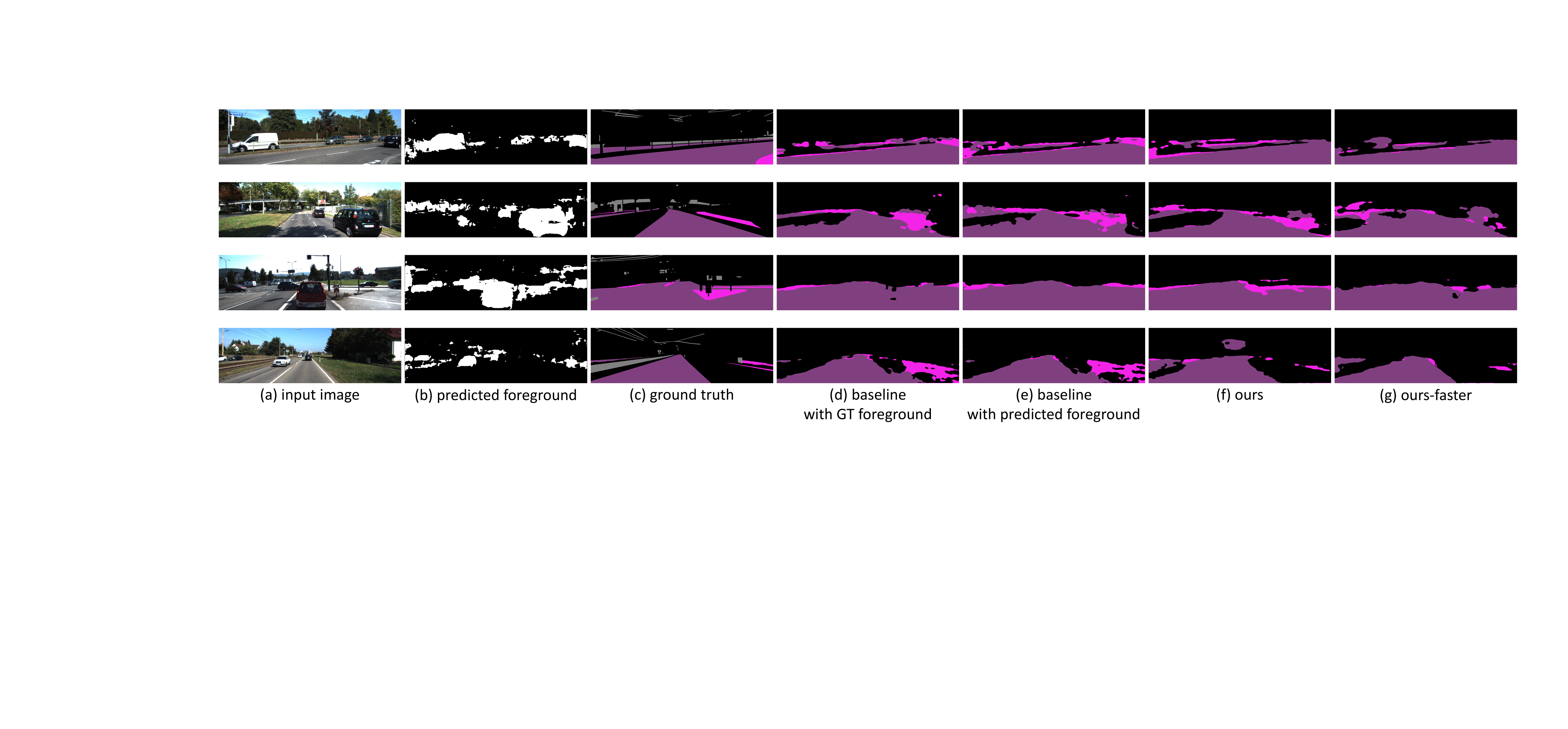}
	\vspace{-10pt}
	\caption{Qualitative results of different methods, which are trained on the Cityscapes dataset, and  tested on the KITTI semantic dataset. (a) is the input RGB image, (b) is the predicted foreground mask by our network, (c) is the manually annotated segmentation of the background, (d)-(g) are the results from the tested methods. Note that the performance is degraded due to the domain gap between Cityscapes and KITTI datasets.}
	\label{fig_kitti_results}
	\vspace{-10pt}
\end{figure*}

On top of this, a second implicit ``supervision" is applied in our method, which is enabled by the usage of our MPI module. After the MPI module, the features at the foreground regions are shared with nearby background features, and the consecutive part of the network is trained to respond to background features by a strong supervision from ground truth labels. Using the MPI module, although the foreground regions are without explicit supervision, the network is able to provide reasonable background predictions at these foreground regions, as the same feature responses are directly learned by the supervision of background segmentation at non-occluded regions. This background feature inpainting and shifting, realized by the MPI module, can be seen as an implicit ``supervision". The ablation study to validate this is provided in Section~\ref{sect_experiments}.

As for the supervision of the network training, two losses are used, namely foreground segmentation loss $\mathcal{L}_{f}$ (binary cross-entropy), and \textit{partial} background segmentation loss $\mathcal{L}_{b}$ (multi-class cross-entropy with the foreground regions ignored). They are formalized as  
\begin{equation}
	\mathcal{L}_{f}(f, \hat{f}) = - \frac{1}{|N|} \sum_{i \in N}  f_i  \log(\hat{f_i}) + (1-f_i) log(1-\hat{f_i}),
\end{equation}
where $N$ is the set of all pixels, $f_i$ and $\hat{f}_i$ denote the ground truth and predicted foreground for each pixel, and
\begin{equation}
	\mathcal{L}_{b}(b, \hat{b}) = - \frac{1}{|A|} \sum_{i \in A} \sum^{C}_{c=1}b_{i, c} \log(\hat{b}_{i, c}),
\end{equation}
where $A$ is the set of pixels at background region, $C$ is the number of background classes, $b_{i, c}$ and $\hat{b}_{i, c}$ denote the ground truth and predicted foreground for each pixel for a certain class. Combining them, the complete loss is
\begin{equation}
	\mathcal{L} = \mathcal{L}_{f} + \mathcal{L}_{b}.
\end{equation}
We do not apply balancing weights on the two losses as they are relatively independent during training.

\section{Experiments}
\label{sect_experiments}

\begin{figure*}[!tbp]
	\centering
	\includegraphics[width=0.95\linewidth]{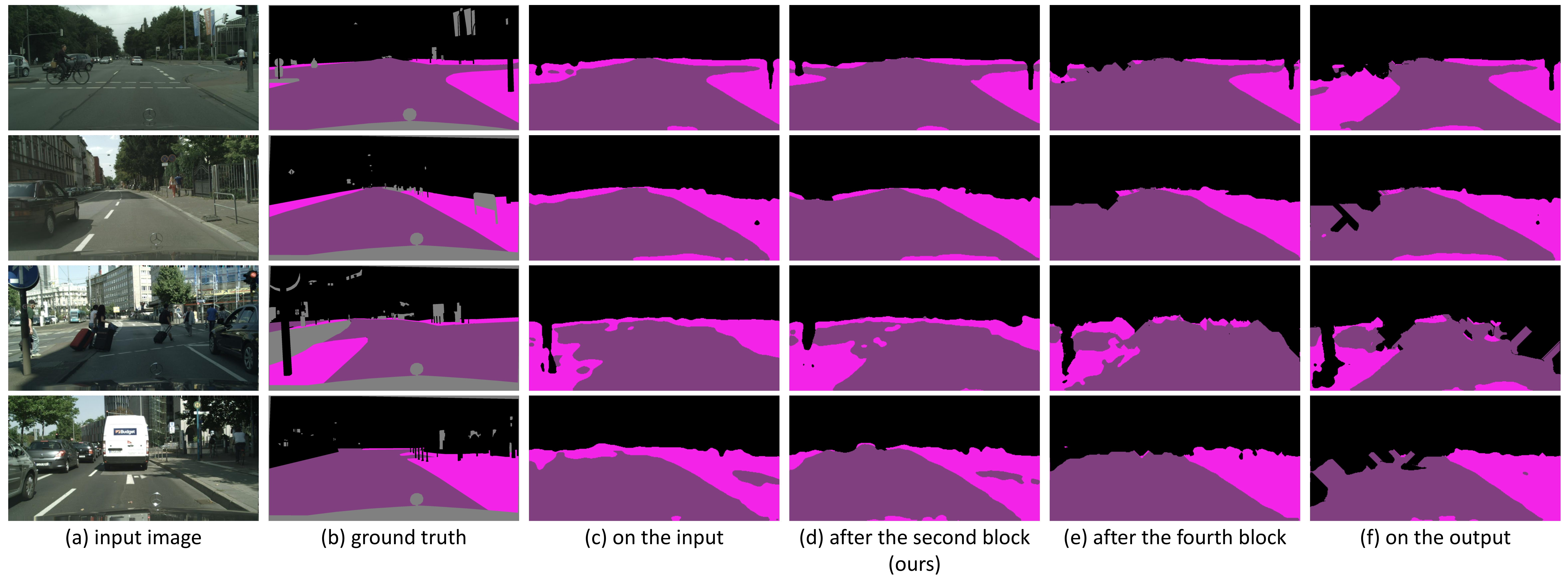}
	\vspace{-10pt}
	\caption{Qualitative comparison of our approaches when the MPI module is applied in the different position of the network. (a) is the input RGB image, (b) is the manually annotated segmentation of the background, (c)-(f) are the cases that MPI is applied on the input RGB image, after the second convolutional block (ours), after the fourth convolutional block, and on the logits output, respectively. }
	\label{fig_different_MPI_position}
	\vspace{-10pt}
\end{figure*}

We perform a series of experiments to illustrate the advantages of the proposed approach for semantic foreground inpainting, as well as the effectiveness of the MPI module. We first introduce the detail settings of the experiments and then discuss the following experimental results:
\begin{itemize}
	\item \textbf{Performance evaluation:} We evaluate the performance and efficiency of our method with that of the existing state-of-the-art approach \cite{schulter_learning_2018}, using the Cityscapes \cite{cordts_cityscapes_2016} dataset.
	\item \textbf{Generalizability evaluation:} To validate the generalizability of our approach, the networks trained on Cityscapes are tested on the KITTI semantics \cite{geiger_vision_2013} dataset, which is not used during training.
	\item \textbf{Ablation studies:} We train the proposed network with different conditions of the MPI module, to validate its claimed functionalities.
\end{itemize}

\subsection{Datasets, training details, and metrics}

\textbf{Datasets:} We use the Cityscapes dataset \cite{cordts_cityscapes_2016} as the primary dataset as it provides sufficient finely annotated segmentation samples. We use the 2975 images in the training set for weakly supervised training. Note that during validation and testing, the ground truth background segmentation at the foreground regions is required. Thus, 500 images in the original Cityscapes validation set are manually annotated at the foreground regions, with the first 100 samples used as the validation set and the remaining 400 samples used as the test set. Furthermore, we use KITTI \cite{geiger_vision_2013} semantic dataset, which contains 200 images with publicly available semantic annotation to verify the generalizability of different methods, with the foreground regions also manually annotated with background classes by us. For the sake of efficiency, if not indicated otherwise, we downsize the images to a height of 256 pixels. Note that in our task, all the classes within the category of ``human" and ``vehicle" are defined as foreground objects. We also simplify the definition of the semantic background to three classes, namely \textit{road}, \textit{sidewalk}, and \textit{other rigid world}. This is because the scene behind the foreground objects is relatively difficult to manually annotate. All the manual annotations are made publicly available to facilitate future research by the community.

\begin{table}
	\renewcommand{\arraystretch}{1}
	\caption{Performance of the proposed approach and the baseline evaluated on the unseen KITTI dataset.}
	\vspace{-10pt}
	\begin{center}
		\begin{tabular}{c||c|c|c|c}
			\hline
			method & backbone & mask used &  IoU(all) & IoU(foreg.) \\
			\hline
			\hline
			baseline \cite{schulter_learning_2018} & \multirow{4}{*}{ResNet-18} & \textit{ground truth} & \textbf{\textit{60.8}} & \textit{40.9}\\
			baseline \cite{schulter_learning_2018} &  & predicted & 57.4 & 37.8 \\
			ours &  & predicted & 59.4 & 39.2 \\
			ours-faster &  & predicted & 59.9 & \textbf{44.1} \\
			\hline
			baseline \cite{schulter_learning_2018} & \multirow{4}{*}{ResNet-50} & \textit{ground truth} & \textit{61.2} & \textit{42.6} \\
			baseline \cite{schulter_learning_2018} &  & predicted & 58.5 & 39.3 \\
			ours &  & predicted & 61.1 & \textbf{45.3} \\
			ours-faster &  & predicted & \textbf{61.5} & 42.6 \\
			\hline
		\end{tabular}
	\end{center}
	\label{tab_exp_kitti}
	\vspace{-15pt}
\end{table}

\textbf{Training details:} We use the same conditions and hyper-parameters as much as possible for the baseline and our method, for a fair comparison. Following the setting of \cite{schulter_learning_2018}, we compare different methods based on the PSPNet \cite{zhao_pyramid_2017} in the main experiments with two backbones, namely ResNet-18 and ResNet-50 \cite{he_deep_2016}. The upsampling module in the main inpainting branch (upper part of Figure \ref{fig_network}) is identical to the one in PSPNet \cite{zhao_pyramid_2017}, while in foreground segmentation branch, the upsampling is performed by bi-linear interpolations and conventional convolutions. As mentioned in~\ref{subsect_semantic_foreground_inpainting} and~\ref{subsect_weak_supervision}, both baseline and our method require a random foreground mask generation process. In the experiments, 3 random rectangular masks are generated for each sample, with the height and width randomly sampled from 0.1 to 0.4 of the corresponding image's (feature map's) height and width. We observe that variance of the number and size of the masks has little effect on the performance. To train the network, we use Adam \cite{kingma_adam:_2014} optimizer with the initial learning rate = 0.0001, $\beta_1$ = 0.6, $\beta_2$ = 0.9, and weight decay = 0.0001. We train both networks for 50 epochs, with batch size 8 (4 when using ResNet-50). The learning rate is decayed by 0.1 for every 20 epochs. All the experiments are performed using Pytorch \cite{paszke_automatic_2017}. For implementation details, our code is made publicly available.

\textbf{Metrics:} Since the prediction is with the same format as semantic segmentation task, we evaluate the results in terms of the commonly used mean intersection-over-union (IoU). Two kinds of mean IoUs are computed for each method: mean IoU of the entire image, and mean IoU of the inpainted foreground regions, which are referred to as \textit{all regions IoU} and \textit{foreground region IoU}, respectively. We take the foreground region IoU as the primary metric and the other one is for checking that the network does not degrade performance for the background. The recorded performances are averaged over the last five training epochs to eliminate small fluctuations. To evaluate the efficiency, we also present, for each method, the \textit{number of parameters} and the \textit{inference time} during testing. The inference time measures the time for a single forward pass of a model with batch size being 1, on the same GPU (NVIDIA Titan V).

\subsection{Performance evaluation}

The results of the main experiment are presented in Table~\ref{tab_main_results}, with the qualitative results shown in Figure~\ref{fig_main_results}. For each backbone network, we present the results of our approaches with or without sharing the early stage features (see two configurations in Figure \ref{fig_network}), which are referred to as \textit{ours} and \textit{ours-faster}, respectively. They are compared with the baseline \cite{schulter_learning_2018} that blacks out the foreground objects region on the input images using two different masks: the ground truth foreground mask, and the predicted foreground mask. The predicted foreground masks are generated by our auxiliary network for a fair comparison.

\begin{table}
	\renewcommand{\arraystretch}{1}
	\caption{Performance of the proposed approach with the MPI module inserted at different positions.}
	\vspace{-10pt}
	\begin{center}
		\begin{tabular}{l||c|c}
			\hline
			MPI position &  IoU (all) & IoU (foreground) \\
			\hline
			\hline
			on the input & 81.7 & 58.1 \\
			after the second block (ours) & \textbf{82.6} & \textbf{60.3} \\
			after the fourth block & 82.5 & 59.0 \\
			on the output & 82.1 & 56.1 \\
			\hline
		\end{tabular}
	\end{center}
	\label{tab_exp_MPIposition}
	\vspace{-15pt}
\end{table}

\begin{table*}
	\renewcommand{\arraystretch}{1}
	\caption{Performance of the proposed approach with and without the MPI module.}
	\vspace{-10pt}
	\begin{center}
		\begin{tabular}{c|c||c|c|c|c}
			\hline
			\multirow{2}{*}{input image size}& \multirow{2}{*}{additional random masks} & \multicolumn{2}{c|}{ with MPI} & \multicolumn{2}{c}{ without MPI}\\ 
			\cline{3-6}
			& &  IoU (all regions) & IoU (foreground region) &  IoU (all regions) & IoU (foreground region) \\
			\hline
			\hline
			256*512 & off & \textbf{81.8} & \textbf{57.1} &  -0.1 (81.7) &  -1.8 (55.3)\\
			256*512 & on & 82.6 & \textbf{60.3} & +0.1 (\textbf{82.7}) & -0.2 (60.1)\\
			384*768 & on & \textbf{84.0} & \textbf{61.1} & -0.4 (83.6) & -2.3 (58.8)\\
			512*1024 & on & \textbf{84.5} & \textbf{61.2} & -0.5 (84.0)& -2.7 (58.5)\\
			\hline
		\end{tabular}
	\end{center}
	\label{tab_ablation}
	\vspace{-10pt}
\end{table*}

Using either backbone network, our approaches outperform the baseline by a margin in both metrics. Compared with the baseline with predicted ground foreground masks, our approach (without sharing the early stage features) is 2.1\% and 3.0\% better at the foreground regions, using ResNet-18 and ResNet-50, respectively. It is worth to note that, even when using the ground truth foreground masks, the baseline is still inferior to our approach that uses the self-contained predicted foreground masks. When sharing the early stage features, the performance of ours-faster slightly degrades (by 0.5\% foreground region IoU) using ResNet-18, and remains the same (with margin less than 0.1\%) using ResNet-50. This is expected, as a strong ResNet-50 backbone can provide adequate intermediate features for both tasks simultaneously, while ResNet-18 has less capacity. 

Besides the quantitative performance, it must be said that our approach is inherently more efficient, in terms of the number of parameters and inference time, as presented in Table~\ref{tab_main_results}. The baseline is a two-stage sequential system as it requires a predicted foreground mask at the input. However in our case, the predicted foreground mask is self-contained and can be computed in parallel with the main branch of the network, which results in a single-stage pipeline.  Evidently, when sharing the early stage features (ours-faster), the efficiency is further boosted especially in terms of the inference time, since a lightweight feature extractor is no longer needed.

\subsection{Generalizability evaluation}

Using the previously trained networks, we feed the unseen KITTI images and evaluate the performance of the two approaches. The results are presented in Table~\ref{tab_exp_kitti} and Figure~\ref{fig_kitti_results}. All the approaches are able to provide acceptable results, with the absolute values of both metrics degraded. This is expected, due to the domain gap between Cityscapes \cite{cordts_cityscapes_2016} and KITTI \cite{geiger_vision_2013}, as they are different in terms of exposure conditions, camera's viewpoints, weather conditions, \etc.

Overall, a similar conclusion can be drawn from Table~\ref{tab_exp_kitti}. Under a fair comparison, \ie, baseline with predicted masks \vs ours(-faster), we outperform the baseline by a large margin in every metric. The same conclusion holds even when the baseline uses the ground truth foreground segmentation as inputs, which is an unfair comparison, especially at the foreground regions. This shows the generalizability of our approach. We notice that the performances are not consistent of ours and ours-faster approaches at foreground regions, when using different backbones: the foreground region IoU drops of ours-faster when using ResNet-50. This is because the first two blocks of ResNet-50 have larger capacity than ResNet-18 for the \textit{foreground segmentation} task, which makes the model over-fitted to Cityscapes and results in worse foreground segmentation quality on KITTI. Binary foreground IoU is 53.2\% of ours-faster with ResNet-50, compared to 61.0\% of ours-faster with ResNet-18. Since the MPI module relies on the quality of foreground segmentation, this explains the inconsistent performance and shows the risk of generalizability when using high capacity backbones.

\begin{figure}[!tbp]
	\centering
	\includegraphics[width=0.95\linewidth]{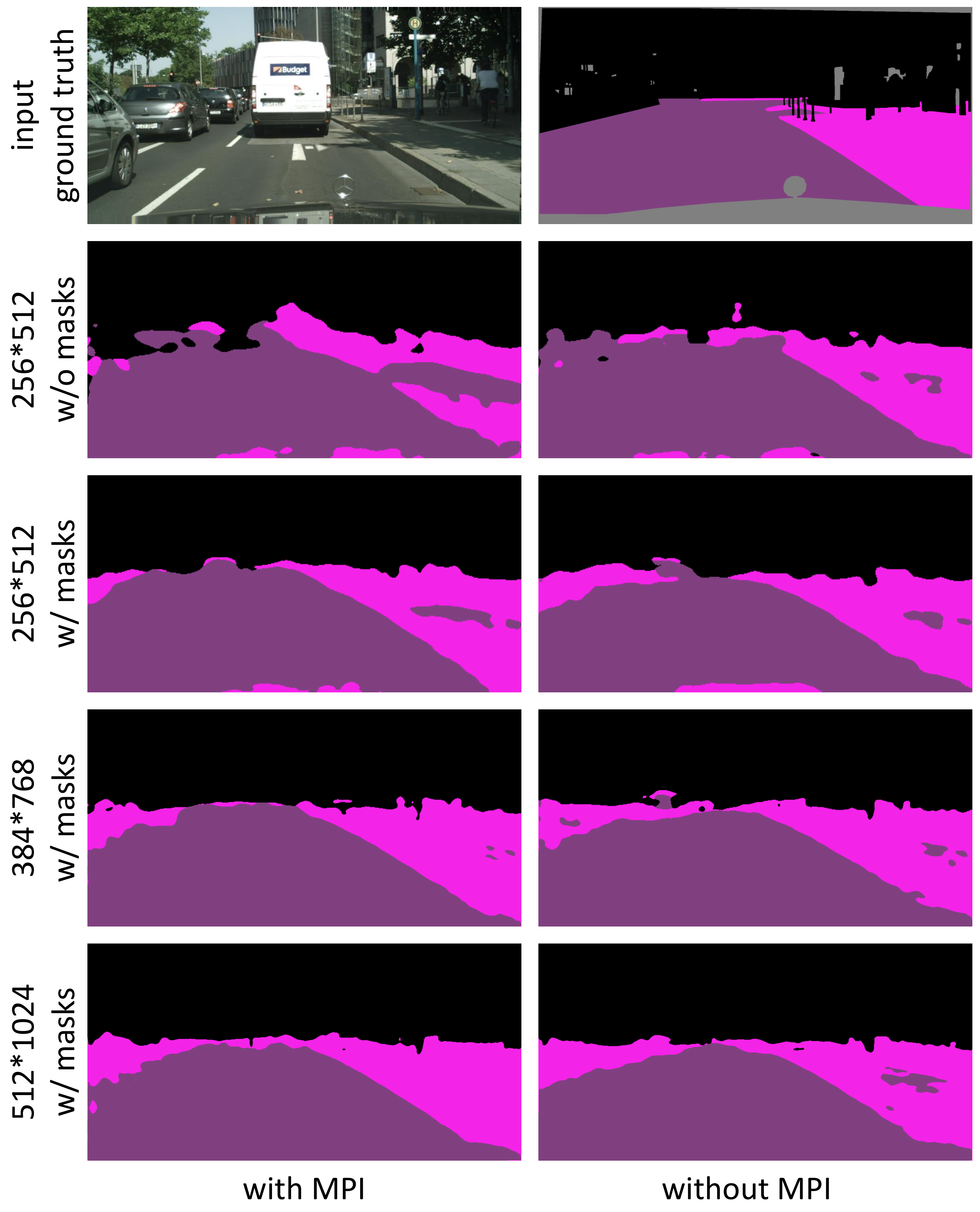}
	\vspace{-10pt}
	\caption{Visualized comparison of our approach conditioned on weather the MPI module is enabled. Different image resolutions and the effect of applying additional random masks are tested. Only one sample is presented due to the space limitation. Quantitative results are listed in Table~\ref{tab_ablation}. }
	\label{fig_ablation}
	\vspace{-10pt}
\end{figure}

\subsection{Ablation studies}

We perform more comprehensive experiments to further validate the proposed MPI module.

\textbf{Where to insert the MPI module:} As mentioned previously, the MPI module can be inserted into a CNN at an arbitrary position. Thus, it is worth to investigate the effect when changing its applied position. In this ablation study, four different positions are tested: 1) on the input images, 2) after the second block of the ResNet feature extractor (ours), 3) after the fourth block of the ResNet feature extractor, and 4) on the final logits predictions. Please note that the position change of the MPI module is non-trivial, especially when applying the MPI on the input image or final logits predictions. Approach 1) can be seen as an entirely new baseline method that applies traditional nearest-neighbor image inpainting as pre-processing, and then trains a canonical CNN for the semantic foreground inpainting task. Also approach 4) can be seen as another baseline method that applies the nearest-neighbor inpainting as post-processing after a canonical semantic segmentation CNN. To isolate the impact of the MPI module, we use ground truth foreground masks in this experiment.

The quantitative results are presented in Table~\ref{tab_exp_MPIposition} with qualitative results in Figure~\ref{fig_different_MPI_position}. Compared with the method that applies the MPI module after the fourth block (right before the PSP module and up-sampling), our approach exhibits better performance in terms of foreground region IoU by 1.3\%, and all regions IoU by 0.1\%. As for the other two approaches that perform the nearest-neighbor inpainting on the input images or logits predictions, we show that our approach surpasses them even further: the margin is 0.9\% and 0.5\% for all regions IoU and 2.2\% and 4.2\% for foreground region IoU, respectively. In conclusion, our setting, which applies the MPI module after the second block of the ResNet backbone, achieves the optimal overall performance.

\textbf{Max-pooling as inpainting v.s. naive blacking out:} To investigate the effect of the MPI module, we perform the control experiments on the functionality of the MPI module. When disabling the MPI module, instead of inpainting the feature maps at the foreground regions, we simply black out these regions (set all foreground feature values to zero). Three image resolutions are tested using the ResNet-18 backbone. Moreover, we also compare two cases when the additional random rectangular masks are disabled. All the other conditions are the same, and in this experiment we also use the ground truth foreground masks to eliminate the effect of foreground prediction quality. The results are listed in Table~\ref{tab_ablation} with some visualized samples in Figure~\ref{fig_ablation}. Please note that it is meaningless to compare the absolute performance between different resolution settings, as the task difficulty and training batch size are different. The performance differences under the same conditions, which are indicated in the brackets, however, are worth to investigate.

When the input image is relatively small, \eg\ 256*512, and the additional random masks are enabled, using the MPI module has little improvements on the performance, which is expected and explainable. The size of the processed feature map is downsized by 8 times (32*64) due to the previous convolutional blocks. In this case, the foreground regions often have a size of three to five pixels. With additional random masks provided during training as a weak supervision, the network is able to learn to fill the foreground regions, which diminishes the difference between the MPI module and a naive blacking out. However, when the additional random masks are disabled, a significant performance degradation (-1.8\% foreground region IoU) can be observed, when not using the MPI module. This indicates that, as mentioned in Section~\ref{subsect_weak_supervision}, the MPI module provides extra weak supervisions to the foreground regions, as it moves the background features, which are trained with the ground truth, into the foreground regions. This facilitates the network's response at the foreground regions and can be seen as implicit supervision.

When the input image is larger, we can see that enabling the additional random masks and disabling the MPI module, results in significantly worse results: -2.3\% for foreground region IoU with image size 384*768, and - 2.7\% with image size 512*1024. This further validates the effectiveness of the MPI module, especially when the images and feature maps are relatively large.

\section{Conclusion}
\label{sect_conclusion}

We presented a CNN architecture that is able to simultaneously predict a foreground and background semantic map, in which regions occluded by foreground objects are inpainted with background classes. This architecture is inherently more efficient than the sequential two-stage state-of-the-art method, and exhibits better semantic inpainting quality. Our approach is enabled by our novel \textit{max-pooling as inpainting} module, which makes use of an efficient canonical max-pooling operation in an iterative manner. The MPI module inpaints the intermediate feature map at the foreground regions using the nearby background features, which boosts the inpainting quality without blocking the gradient flow during network training. In future work, we plan to expand the semantic foreground inpainting task to other scenarios such as indoor images, and showcase how the improved semantic inpainting could facilitate robotic applications such as semantic mapping.

\bibliographystyle{./IEEEtran}
\bibliography{./MyLibrary.bib}

\end{document}